%% file: main.tex
\documentclass[runningheads]{llncs}

\usepackage{eccv}
\usepackage{fancyhdr}

\usepackage{eccvabbrv}

\usepackage{graphicx}
\usepackage{booktabs}
\input{packagelist}

\usepackage[accsupp]{axessibility}  

\usepackage[pagebackref,breaklinks,colorlinks,citecolor=eccvblue]{hyperref}

\usepackage{orcidlink}

\begin{document}

\title{Deep Feature Surgery: Towards Accurate and Efficient Multi-Exit Networks}


\author{Cheng Gong\inst{1}\orcidlink{0000-0002-6594-8375} \and
	Yao Chen\inst{3*} \and
	Qiuyang Luo\inst{1} \and
	Ye Lu\inst{2,6,7}\orcidlink{0000-0003-0805-6394} \and
	Tao Li\inst{2,5} \and
	Yuzhi Zhang\inst{1,5} \and
	Yufei Sun\inst{1*,5} \and
	Le Zhang\inst{4}
}

\authorrunning{Cheng Gong et al.}

\institute{College of Software, Nankai University\\
	\email{yufei\_sun@sina.com}
	\and
	College of Computer Science, Nankai University \and
	National University of Singapore\\
	\email{yaochen@nus.edu.sg}
	\and
	School of Information and Communication Engineering, University of Electronic Science and Technology of China \and
	HAIHE Lab of ITAI \and
	Tianjin Key Laboratory of Network and Data Security Technology \and
	Key Laboratory of Data and Intelligent System Security, Ministry of Education
}

\maketitle
\thispagestyle{fancy}
\fancyhead{}
\lhead{Accepted as a conference paper in ECCV 2024}
\renewcommand{\headrulewidth}{1pt}

\renewcommand{\thefootnote}{}
\footnotetext{{* Corresponding Author}}
\input{sections/00-Abstract.tex}
\input{sections/01-Introduction.tex}
\input{sections/02-Relateworks.tex}
\input{sections/03-dfs.tex}

\input{sections/03-Method.tex}
\input{sections/04-Implementation.tex}
\input{sections/05-Experiments.tex}

\input{sections/06-Conclusion.tex}
\input{sections/07-Acknowledge}

%
%
\bibliographystyle{splncs04}
\bibliography{references}
\end{document}

%% file: packagelist.tex
\usepackage{amssymb}
\usepackage{latexsym}
\usepackage{url}
\usepackage{xcolor}
\definecolor{newcolor}{rgb}{.8,.349,.1}

\usepackage{epsfig}
\usepackage{caption}
\usepackage{comment}
\usepackage{amsmath}
\usepackage{cases}
\usepackage{esint}
\usepackage{graphicx}
\usepackage{wrapfig}
\usepackage{algorithm} 
\usepackage{algorithmic} 
\usepackage{multirow} 
\usepackage{xcolor}

\usepackage{overpic}
\newcommand{\secref}[1]{Sec. \ref{#1}}
\newcommand{\figref}[1]{Fig. \ref{#1}}

\newcommand{\tabref}[1]{Tab. \ref{#1}}

\newcommand{\methodname}[0]{DFS}

%% file: sections/00-Abstract.tex
\begin{abstract}
	Multi-exit network is a promising architecture for efficient model inference by sharing backbone networks and weights among multiple exits.
	However, the gradient conflict of the shared weights results in sub-optimal accuracy.
	This paper introduces Deep Feature Surgery (\methodname),
	which consists of feature partitioning and feature referencing approaches to resolve gradient conflict issues during the training of multi-exit networks.
	The feature partitioning separates shared features along the depth axis among all exits to alleviate gradient conflict while simultaneously promoting joint optimization for each exit.
	Subsequently, feature referencing enhances
	multi-scale features for distinct exits across varying depths to improve the model accuracy.
	Furthermore, \methodname~reduces the training operations with the reduced complexity of backpropagation.
	Experimental results on Cifar100 and ImageNet datasets exhibit that \methodname~provides up to a \textbf{50.00\%} reduction in training time and attains up to a \textbf{6.94\%} enhancement in accuracy when contrasted with baseline methods across diverse models and tasks.
	Budgeted batch classification evaluation on MSDNet demonstrates that DFS uses about $\mathbf{2}\boldsymbol{\times}$ fewer average FLOPs per image to achieve the same classification accuracy as baseline methods on Cifar100.
	The code is available at \url{https://github.com/GongCheng1919/dfs}.
\end{abstract}

%% file: sections/01-Introduction.tex
\section{Introduction}

A multi-exit network refers to a neural network architecture designed with multiple points at which the network outputs a result or makes a decision, rather than having a single endpoint.
These intermediate exits can be utilized for various purposes such as early exiting for efficiency, where the network can provide a quick prediction without the need for full computation through the entire network depth.
Multi-exit networks are widely applied in domains including dynamic computing graph~\cite{han2022dynamic}, anytime prediction~\cite{li2019improved}, and collaborative inference~\cite{lin2019computationoffloading}.

In multi-exit network training, it is imperative to formulate multiple losses for distinct exits and concurrently optimize these losses.
However, multiple exits manifest inconsistent or diametrically opposing gradient directions, denoted as the gradient conflict issue~\cite{chai2023improving,deng2023split,hu2023gradient,
    yu2020gradient,chen2018gradnorm,liu2021conflict,
    sun2023learning}.
This issue hinders the convergence of multi-exit networks, consequently exerting a
impact on the model accuracy.

\begin{figure*}[!t]
    \centering
    \includegraphics[width=1\textwidth]{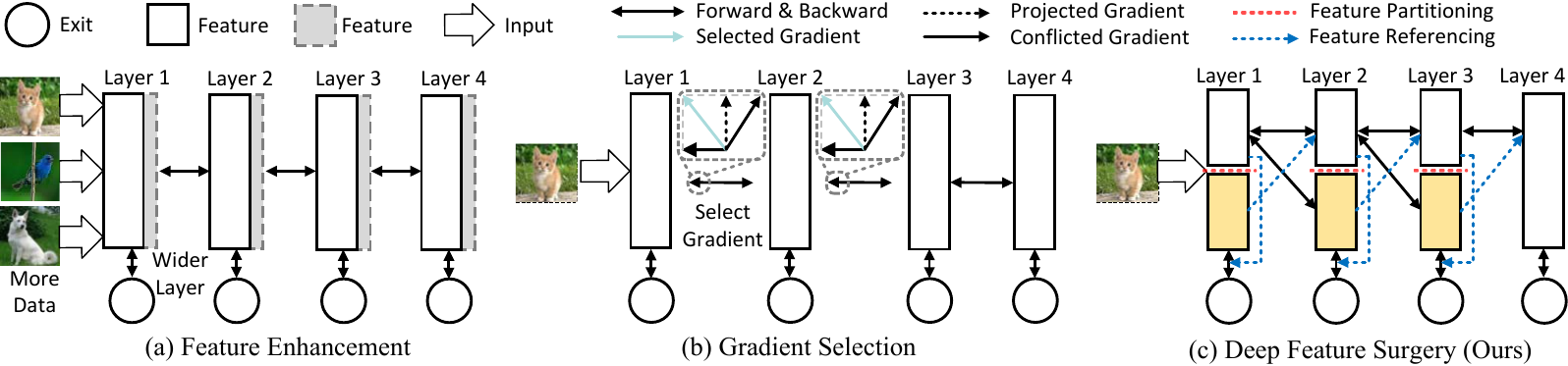}
    \vspace{-15pt}
    \caption{
        Solutions to improve the accuracy of the exits for multi-exit networks.
        (a) Feature enhancement introduces additional features to improve the accuracy of exits.
        (b) Gradient selection adjusts gradients from conflicting sources to attain consistent update directions.
        (c) Our method alleviates gradient conflict and ensures end-to-end joint optimization of exits.
    }
    \label{fig:feature_grouping}
    \vspace{-20pt}
\end{figure*}

Recent research advancements have introduced enhancements for training multi-exit networks, primarily categorized into feature enhancement~\cite{li2019improved,zhang2019your,zhang2019scan,phuong2019distillation,han2022learning} and gradient selection~\cite{li2019improved,sun2022meta,hu2023gradient}.
Feature enhancement introduces additional features (more data or wider layer) and training iterations to improve the accuracy of the exits of the model as shown in \figref{fig:feature_grouping}\textcolor{red}{a}.
This approach is irrelated to the gradient conflict issue and may lead to elevated training expenses and necessitate model redesign~\cite{li2019improved,zhang2019your,zhang2019scan,phuong2019distillation,han2022learning}.
Gradient selection adjusts gradients from conflicting sources to attain consistent update directions conducive to model optimization as shown in \figref{fig:feature_grouping}\textcolor{red}{b}.
This approach is generally heuristic, demonstrating effectiveness in specific models but facing accuracy challenges for universal adoption~\cite{li2019improved,sun2022meta,hu2023gradient}.
In addition, it has to take additional forward-backward phases to merge the gradients
to adjust the update directions of the parameters, leading to a long model training time~\cite{liu2021conflict,yu2020gradient,navon2022multi,sun2022meta}.

This paper proposes Deep Feature Surgery (\methodname) to tackle both the accuracy and training efficiency challenges in training multi-exit networks.
Specifically, \methodname~utilizes two techniques for feature tensors to address the gradient conflict and enables joint optimization to the different groups of parameters, namely feature partitioning and feature referencing.
Feature partitioning divides the features of different exits along the depth axis into exit-specific and shared components as shown in \figref{fig:feature_grouping}\textcolor{red}{c}, and updates the corresponding parameters only during the backpropagation.
This partitioning reduces the shared features and the corresponding shared parameters among distinct exits, effectively alleviating gradient conflict while ensuring end-to-end joint optimization of exits with varying depths.
Feature referencing allows exits with different depths to reuse the
exit-specific features to enhance their accuracy.
In addition, \methodname~reduces the back-propagation operations and improves the model training efficiency by reducing a substantial portion of gradient computation during model training.

In summary, our contributions are as follows:
\begin{itemize}
    \item We propose \methodname, which consists of feature partitioning and feature referencing to address the gradient conflict issue of training multi-exit networks.
    \item We reduce the required operations during model training with \methodname~to improve the training efficiency of multi-exit networks.
    \item Extensive experiments demonstrate that \methodname~consistently outperforms baseline methods across various models and tasks with up to $\textbf{50.00\%}$ reduction in training time and a maximum improvement of $\textbf{6.94\%}$ in accuracy.
\end{itemize}


%% file: sections/02-Relateworks.tex
\section{Related work}
Training multi-exit networks is challenging since the gradients from multiple exits hold conflicts, which usually lead to sub-optimal model accuracy~\cite{yu2020gradient}.

\subsection{Feature Enhancement}

Feature enhancement tries to introduce more robust features to improve the accuracy of multi-exit networks which includes knowledge transfer~\cite{li2019improved}, knowledge distillation~\cite{li2019improved,zhang2019your,zhang2019scan,phuong2019distillation}, weighted sample learning~\cite{han2022learning}, and dense connections~\cite{huang2018MSDNet}.
Since feature enhancement is an irrelative technique for solving gradient conflict issues, it was combined with other techniques to further improve the performance of multi-exit architectures.
Feature enhancement methods often necessitate extensive modifications to the model~\cite{huang2018MSDNet}, loss function~\cite{li2019improved,zhang2019your}, and even the dataset~\cite{han2022learning}, thereby impacting the model training efficiency.

\subsection{Gradient Selection}

Gradient selection is effective in many applications such as computer vision~\cite{liu2021conflict,sun2022meta} and natural language processing~\cite{hu2023gradient,chai2023improving}
by selecting a proper updating direction.
One way is to adjust the gradients to multiple exits.
For example, in \cite{li2019improved}, researchers develop a gradient equilibrium strategy to reduce the gradient variance and stabilize the training process; Meta-GF~\cite{sun2022meta} take account of the importance of the shared weights to each exit, and introduce a meta-learning-based method for weighted fusion of the gradients of each exit; \cite{hu2023gradient} propose Gradient Remedy (GR) to change the angle and magnitude of gradient vectors to solve interference between two gradients in noise-robust speech recognition.
Another way is to balance the multiple losses to the multiple exits and obtain better-weighted gradients without complex intervention. \cite{guo2018dynamic} design an adaptive loss-weighting policy to prioritize more difficult tasks. GradNorm~\cite{chen2018gradnorm} and \cite{liu2021towards} introduce the magnitude re-scaling algorithms to balance the gradient magnitudes of different tasks.
Other widely used gradient selection methods are multi-objective algorithms, such as the MGDA~\cite{sener2018MGDA}, PCGrad~\cite{yu2020gradient}, CAGrad~\cite{liu2021conflict}, GetMTL~\cite{chai2023improving}, Nash-MTL~\cite{navon2022multi}, and so on.
These methods formulate multi-exit network training as multi-objective optimization and try to pursue the Pareto optimalities for better trade-offs among multiple losses to multiple exits.
Gradient selection inevitably degrades the overall accuracy of networks in the gradient adjusting and compromises among multiple exits, \ie, for one loss to give up updating others.
Furthermore, this technique requires computing independent weight gradients for each exit to calculate the merge coefficients,
which often entails multiple forward and backward passes for one training iteration, resulting in a manifold increase in training time.

\subsection{Feature partitioning}
Feature partitioning is a technique proposed for mitigating task interference in multi-task learning~\cite{ding2023mitigating, strezoski2019many, pascal2021maximum, maninis2019attentive, liu2019end, deng2023split}.
For instance,
ETR-NLP~\cite{ding2023mitigating}
partitions the features into task-specific and task-shared ones, to update the corresponding parameters regarding different tasks during model training.
It trains the task-specific and shared parameters separately, thus largely increasing training iterations and delaying model training.
Mask-based works~\cite{strezoski2019many,pascal2021maximum} learn the mask module for each task during training to select the proper parameters.
The mask modules for different tasks are optimized separately which requires a long training time.
\cite{maninis2019attentive,liu2019end} employ the attention mechanisms at the filter level as feature selectors allowing each task to select a subset of parameters.
They train the attention modules and model parameters simultaneously in one training iteration where attention modules delay the training.
GradSplit~\cite{deng2023split} splits features into multiple groups for multiple tasks and trains each group only using the gradients back-propagated from the task losses.
These studies have demonstrated success in multi-task learning, but they do not translate effectively to training multi-exit networks due to the training efficiency issue and gap between multi-task learning and multi-exit networks.
To the best of our knowledge, DFS is the first study that employs feature partitioning to mitigate the gradient conflict issue in training multi-exit networks, and significantly enhances both model accuracy and training efficiency of multi-exit networks.

%% file: sections/03-dfs.tex
\section{Preliminary: Gradient Conflict}
Gradient conflict is the major cause of the accuracy degradation in multi-exit network training.
Let a feature extraction architecture, such as a convolutional neural network, with $L$ layers be the backbone network.
Take the classification task as an example,
this multi-exit network has $L$ tasks for the exits, denoted as $C=\{c_1,c_2,\cdots,c_L\}$ and weights for these layers from the backbone network, denoted as $W_L=\{w_1,w_2,\cdots,w_L\}$.
Since each exit also has its own exit-specific weights, denoted as $W_C=\{w_{c_1},w_{c_2},\cdots,w_{c_L}\}$.
Let $(X, Y)$ be a dataset, in which $X$ is the sample set and $Y$ is the labels of $X$, and the $i-$th task $c_i$ of the given multi-exit network is defined as follows.
\begin{equation}\label{eq:mcn_forward}
    \small
    c_i=\pi_i(f_i,w_{c_i}), ~~s.t.
    \begin{cases}
        f_i=\varphi_i(X,W_i), \\
        i=1,2,\cdots, L
    \end{cases}
\end{equation}
Here, $\pi_i(\cdot)$ processes the exit-specific output with the exit-specific weight $w_{c_i}$ and the intermediate feature $f_i$ from the $i$-th layer of backbone network.
$f_i$
is extracted by $\varphi_i(\cdot)$ with input $X$ and weights $W_i$.
For weight sets $W_i$ and $W_j$, they share weights as follows.
\begin{equation}\label{eq:shared_weights}
    \small
    W_i \cap W_j = \{w_1,w_2,\cdots,w_{\min(i,j)}\},
    ~~~s.t.
    \begin{cases}
        i=1,2,\cdots, L \\
        j=1,2,\cdots, L
    \end{cases}
\end{equation}
As a result, the optimization formulation of this multi-exit network can be derived as follows.
\begin{equation}\label{eq:multi-optimization}
    \small
    W^*_L,W_C^*=\underset{W_L,W_C}{\arg\min}\{CE(c_1,Y),CE(c_2,Y),~\cdots,CE(c_L,Y)\}.
\end{equation}
Here, $CE(\cdot)$ computes the cross entropy of prediction $\{c_i|i=1,2,\cdots,L\}$ and category $Y$.
Obviously, \cref{eq:multi-optimization} is a multi-objective optimization~\cite{sener2018MGDA} problem, which implies that it is hard to achieve the optimal solutions for all classification tasks simultaneously.

The most common methods employ a gradient descent algorithm to optimize \cref{eq:multi-optimization}, the gradients $g_{w_i}$ of tasks with respect to weight $w_i$ is computed as the summation of $\{g^{k}_{w_i}|k=i,i+1,\cdots,L\}$ as follows since $w_i$ is the shared weight among these exits~\cite{huang2018MSDNet}.
\begin{equation}\label{eq:src_constraints}
    \small
        g_{w_i} = \sum_{k=i}^{L}g^{k}_{w_i}, ~~
        s.t.~~g^{k}_{w_i}=\frac{\partial CE(c_k,Y)}{\partial w_i}
\end{equation}
Although $g^k_{w_i}$ implies the optimal updating direction of $w_i$ to reduce $CE(c_k, Y)$.
The inconsistency in the gradients of $\{g^k_{w_i}|k=1,2,\cdots,L\}$ is known as gradient conflict and affects the training performance.

%% file: sections/03-Method.tex
\section{Deep Feature Surgery}

\begin{figure*}[!t]
    \centering
    \includegraphics[width=1\textwidth]{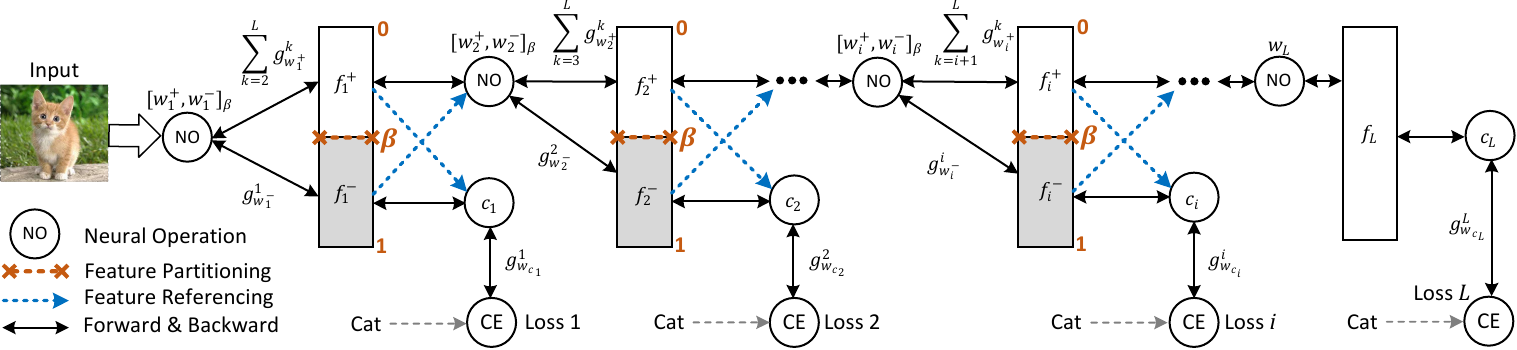}
    \vspace{-15pt}
    \caption{
        Deep feature surgery for multi-exit networks. The black arrow represents forward and backward propagation. The blue arrow indicates feature reference. The red-doted line represents the partitioning position. DFS splits the features of each layer into two distinct parts $f_i^+$ and $f_i^-$ with coefficient $\beta$, which reduces the number of shared weights $w_i^+$ among different exits thus mitigating gradient conflict and reducing backward computation operations.
        DFS cross-references the shared features and exit-specific features among exits with varying depths in the forward phase while ignoring this in the backward phase, thus using more features for predicting tasks while not introducing more inconsistent gradients.
    }
    \label{fig:dynamic_neural_surgery_framework}
    \vspace{-20pt}
\end{figure*}

\methodname~addresses the gradient conflict by creating both individual and shared parameter updates.
\methodname~conducts feature partitioning and feature referencing on the feature tensors to ensure the desirable model accuracy of different exits.

\subsection{Feature Partitioning}\label{sec:feature_partitioning}
Gradient conflict is caused by the shared weights of exits in multi-exit networks, as shown in \cref{eq:shared_weights}.
In order to alleviate the gradient conflict, 
we first decouple the weight $w_i$ of $i$-th layer into two distinct parts with a coefficient $\beta$. 
Let $w_i\in \mathbb{R}^{C_{in}\times C_{out}\times K\times K}$,
we define:
\begin{equation}\label{eq:weight_partitioning}
    \small
    w_i=[w_i^+,w_i^-]_{\beta}, ~s.t.
    \begin{cases}
        w_i^+ = w_i[:,:\lfloor\beta C_{out}\rceil,:,:] \\
        w_i^- = w_i[:,\lfloor\beta C_{out}\rceil:,:,:]
    \end{cases}
\end{equation}
Here, $\lfloor\cdot\rceil$ computes the rounding result of input, and $w_i[:]$ represents a slicing operation on the tensor $w_i$.
$[\cdot]_\beta$ is the concatenate operation of tensors. $\beta\in (0,1)$ indicates the partitioning ratio of $w_i^+$ and $1-\beta$ is that of $w_i^-$.
$w_i^+$ is shared among exits and $w_i^-$ is exit-specific weights.
Based on the partitioned weights, task $c_i$ is computed as follows.
\begin{equation}\label{eq:dividied-tasks}
    \small
        c_i=\pi_i(f_i^-,w_{c_i}),
        ~~~s.t.
        \begin{cases}
            f_i^-=\varphi_i(X,\{w_1^+,w_2^+,\cdots,w_{i-1}^+,w_i^-\}) \\
            i=1,2,\cdots,L-1.
        \end{cases}
\end{equation}
According to the optimization in \cref{eq:multi-optimization}, the gradients $g_{w_i}$ of $w_i$ can be computed as follows.
\begin{equation}\label{eq:alleviate_gradient_conflicts}
    \small
    g_{w_i}=[\sum_{k=i+1}^L g_{w_i^+}^k,g^i_{w_i^-}]_{\beta},~~
    s.t.~g_{w_i^+}^k=\frac{\partial CE(c_k,Y)}{\partial
        w_i^+},~g^i_{w_i^-}=\frac{\partial CE(c_i,Y)}{\partial w_i^-}.
\end{equation}

Through decoupling weights into two distinct parts, we alleviate the gradient conflict of $\sum_{k=i}^{L}g^{k}_{w_i}$ in \cref{eq:src_constraints} to $\sum_{k=i+1}^{L}g^{k}_{w_i^+}$ in \cref{eq:alleviate_gradient_conflicts} as shown in \figref{fig:dynamic_neural_surgery_framework}.
Notably, $\beta\rightarrow 1$ partitions all elements in $w_i$ into $w_i^+$,
the task $c_i$ is invalid.
On the contrary, $\beta\rightarrow 0$ partitions all elements in $w_i$ into $w_i^-$ leads to $|w_i^+|\rightarrow 0$,
the multi-exit network degrades into a cascade of network layers, and the end-to-end joint optimization from input $X$ to tasks is eliminated.
We constrain the value of coefficient $\beta$ to the range of $(0,1)$ to alleviate the gradient conflict in multi-exit network optimization while ensuring the joint optimization of tasks.

Instead of partitioning the weight tensor, we partition the features for easy deployment in practice.
As shown in \figref{fig:dynamic_neural_surgery_framework}, taking the first layer as an example, the output feature $f_1=\varphi_1(X,w_1)$ of the first layer is split into $f_1^+=\varphi_1(X,w_1^+)$ and $f_1^-=\varphi_1(X,w_1^-)$, which is corresponding to weight partitioning in \cref{eq:weight_partitioning}.
The corresponding feature partitions to \cref{eq:dividied-tasks} is as follows:
\begin{equation}\label{eq:weightdivide_eq_featuredivide}
    \small
    \begin{cases}
        f_i^-=\varphi_i(X,\{w_1^+,w_2^+,\cdots,w_{i-1}^+,w_i^-\}) \\
        f_i^+=\varphi_i(X,\{w_1^+,w_2^+,\cdots,w_{i-1}^+,w_i^+\})
    \end{cases}, ~~s.t.~i=1,2,\cdots,L-1.
\end{equation}
Here, $f_i^+$ is shared on all the following tasks including $\{c_{i+1},c_{i+2},\cdots,c_L\}$,
and $f_i^-$ is utilized for predicting $c_i$.
$c_i$ is the only task that contributes to learning $w_i^-$ by supervising the exit-specific feature $f_i^-$.

\subsection{Feature Referencing}\label{sec:fr}

Feature partitioning alleviates the gradient conflict in multi-exit network training but reduces the number of features for weight training.
For example, $f_{i}^+$ is not available for task $c_{i}$ in \cref{eq:dividied-tasks}.
The reduced number of features, \eg, $f_i^+$, can easily result in the under-fitting of tasks. 
We introduce feature referencing to address this issue.
Feature referencing reuses these features across tasks to benefit the task performance.
Based on feature referencing, task $c_i$
and gradients $g_{w_i}$ of all tasks with respect to weight $w_i$ are defined as follows.
\begin{equation}\label{eq:fsh}
    \small
    \begin{cases}
        c_i   & =\pi_i([f_i^+,f_i^-]_{\beta},w_{c_i})            \\
        f_i^+ & = \varphi_i([f_{i-1}^+,f_{i-1}^-]_{\beta},w_i^+) \\
        f_i^- & =\varphi_i([f_{i-1}^+,f_{i-1}^-]_{\beta},w_i^-)
    \end{cases},
    ~~s.t.~
    \begin{cases}
        f_0=X, \\
        i=1,2,\cdots,L-1
    \end{cases}
\end{equation}
Here we use both the shared feature $f_{i-1}^+$ and exit-specific feature $f_{i-1}^-$ for extracting next-layer features $f_i^+$ and $f_i^-$ in the forward phase with partitioned weights,
while both $f_i^+$ and $f_i^-$ are used for predicting task $c_i$.
This approach increases the involved number of features from the previous output, thus increasing the number of exit-specific weight $w_{c_i}$ for task $c_i$ to improve the accuracy of the task.
In the backward phase,
to eliminate the gradient conflict, the computation of gradients $\{g_{w_i^-}^k|k=i+1,\cdots,L\}$ is avoided.
As shown in \cref{fig:dynamic_neural_surgery_framework}, for the exit-specific weight $w_i^-$ in $i$-th layer, we only compute the gradients $g_{w_i^-}^i$ from the $i$-th loss to ensure the gradient $g_{w_1^-}$ has no conflicts.

Feature referencing uses more features for predicting task $c_i$ while not introducing more inconsistent gradients for updating weight $w_i$.
Moreover, the feature segments $f_i^+,$ and $f_i^-$ are referenced by different exits with varying depth, and the tasks for these exits supervise the $w_i^+$ and $w_i^-$ to learn varying features with different scales
which improve the accuracy of multi-exit networks \cite{huang2018MSDNet}.

%% file: sections/04-Implementation.tex
\section{Training Efficiency Improvement}\label{sec:implementation}

The training efficiency of multi-exit network models hinders their wide adoption.
\methodname~not only improves the accuracy of the different exits but also improves the training efficiency of multi-exit network models by reducing the total number of training operations.
To simplify the representation,
we assume the neural operations (NO) in \figref{fig:dynamic_neural_surgery_framework} only contain matrix-multiply operations to describe the reduction of Multiply-Accumulation operation in the training of multi-exit networks with \methodname.

Let the $i$-th neural operation of a vanilla multi-exit network be:
\begin{equation}\label{eq:original_forward}
    \small
    \begin{cases}
        f_i=\varphi_i (f_{i-1}\cdot w_i) \\
        l_i=CE(\pi_i(f_i\cdot w_{c_i}),Y)
    \end{cases}, ~s.t.
    \begin{cases}
        f_0=X, \\
        i=1,2\cdots, L.
    \end{cases}
\end{equation}
Here $f_i$ and $f_{i-1}$ are the features for layer $i$ and $w_i$ is the corresponding layer weights.
There are $L$ tasks with loss set $\{l_i|i=1,2,\cdots,L\}$, and task losses $\{l_j|j>=i\}$ reference the same intermediate feature $f_i$, recursively.
For simplicity, we assume the matrixes $f_i$, $w_{i}$, and $w_{c_i}$ have the same shape [$N, N$].
Ignoring the computing operations of activation and loss functions ($\varphi_i(\cdot)$ and $CE(\cdot)$), the operations for computing $f_{i-1}\cdot w_i$ and $f_i\cdot w_{c_i}$ are both $2N^3$, and the total forward operations are $4LN^3$.

We can further derivate the gradients of all task losses with respect to the $w_i$ matrix as follows.
\begin{equation}\label{eq:vallina_backward}
    \small
    \begin{split}
        g_{w_i} &= \frac{\partial l_i}{\partial w_i} + \frac{\partial l_{i+1}}{\partial w_i}+\cdots+\frac{\partial l_L}{\partial w_i}\\
        &=\sum_{k=i}^{L}(\frac{\partial l_k}{\partial f_k}\Pi_{j=i}^{j<k}(\frac{\partial f_{j+1}}{\partial f_{j}}))\cdot w_i^T
        ~s.t.~i=1,2,\cdots,L.
    \end{split}
\end{equation}
Because of the chain rules and back-propagation algorithms~\cite{rumelhart1986BPlearning}, it only requires to compute $L-1$ times ${\partial f_{j+1}}/{\partial f_{j}}$, $L$ times ${\partial l_k}/{\partial f_k}$, and one ${\partial f_i}/{\partial w_i}$ to obtain $g_{w_i}$.
The operations for computing gradients $g_{w_i}$ are $4LN^3$.
Since there are total $2L$ trainable weights ($w_i$ and $w_{c_i}$), and the operations for computing ${\partial f_i}/{\partial w_i}$ and ${\partial l_i}/{\partial w_{c_i}}$ are both $2N^3$ (one matrix multiplication), the total backward operations are $(8L-2)N^3$, and the total forward-backward operations are $(12L-2)N^3$ for one training step for a vanilla multi-exit network.

Assuming that $N$ can be evenly divided by $1/\beta$,
then the $i$-th feature $f_i$ is partitioned into $f_i^+\in \mathbb{R}^{N*(\beta N)}$ and $f_i^-\in \mathbb{R}^{N*{(1-\beta)N}}$, and the neural operations are rewritten as follows:
\begin{equation}\label{eq:dns_forward}
    \small
        f_i = [f_i^+,f_i^-]_{\beta}=[\varphi_i(f_{i-1}\cdot w_1^+),\varphi_i(f_{i-1}\cdot w_1^-)]_{\beta} 
        ~s.t.~f_0=X, i=1,2,\cdots,L.
\end{equation}

During the backward process, as described in \cref{eq:fsh} in Section \ref{sec:fr}, only the gradients from an exit-specific loss $l_i$ are adopted for weight $w_i^-$, which have no conflicts, and all gradients from all losses are adopted to update $w_i^+$ for end-to-end optimization.
So the computation of $g_{w_i}$ in \cref{eq:alleviate_gradient_conflicts} is expanded as follows.

\vspace{-10pt}
\begin{equation}\label{eq:backward_with_dns}
    \small
    \begin{cases}
        g_{w_i^+} & =\sum_{k=i+1}^L {g_{w_i^+}^k}
        ={g_{w_i^+}^L}+\sum_{k=i+1}^{L-1}(\frac{\partial l_{k}}{\partial f_{k}^-}
        \frac{\partial f_{k}^-}{\partial f_{k-1}^+}
        \Pi_{j=i}^{j<k-1}(\frac{\partial f_{j+1}^+}{\partial f_{j}^+}))\frac{\partial f_{i}^+}{\partial w_i^+}                                                    \\
        g_{w_i^-} & =g_{w_i^-}^i=\frac{\partial l_i}{\partial f_i^-}\cdot\frac{\partial f_i^-}{\partial w_i^-}=\frac{\partial l_i}{\partial f_i^-}\cdot(w_i^-)^T.
    \end{cases}
\end{equation}

The feature number of $f_i^+$ and $f_i^-$ is $\beta$ and ($1-\beta$) times that of $f_i$.
Therefore, the operations for computing ${\partial l_i}/{\partial f_i^+}$ and ${\partial l_i}/{\partial f_i^-}$ are also $\beta$ and ($1-\beta$) times that for ${\partial l_i}/{\partial f_i}$, and they are $\beta\cdot2N^3$ and $(1-\beta)\cdot2N^3$, respectively.
The operations to compute ${\partial f_k^-}/{\partial f_{k-1}^+}$ and ${\partial f_{j+1}^+}/{\partial f_j^+}$ are $2\beta(1-\beta)N^3$ and $2\beta^2N^3$, respectively.
Employing the back-propagation algorithm~\cite{rumelhart1986BPlearning}, we only need to compute
$L$ times of ${\partial l_k}/{\partial f_k^-}$, $L-2$ times of ${\partial f_{k}^-}/{\partial f_{k-1}^+}$,
$L-2$ times of ${\partial f_{j+1}^+}/{\partial f_{j}^+}$ and
one time of ${\partial l_{L}}/{\partial f_L}$ and ${\partial f_L}/{\partial f_{L-1}^+}$ to compute $g_{w_i^+}^L$), and one time of ${\partial f_{i}^+}/{\partial w_i^+}$ and ${\partial f_{i}^-}/{\partial w_i^-}$ in \cref{eq:backward_with_dns}.
The operations for computing $g_{w_i}=[g_{w_i^+},g_{w_i^-}]_{\beta}$ is $2(L+1)N^3$, which is irrelated to coefficient $\beta$.
There are a total number of $2L$ trainable weights ($w_i$ and $w_{c_i}$), and the operations for computing $[\frac{\partial f_i^+}{\partial w_i^+},\frac{\partial f_i^-}{\partial w_i^-}]_{\beta}$ and $\frac{\partial l_i}{\partial w_{c_i}}$ are both $2N^3$, the total backward operations are $6LN^3$. 
The total forward-backward operations are $10LN^3$ for one training step with \methodname.
Compared with one vanilla training step, applying \methodname~reduces the operations by up to $16.67\%$, which is formulated as follows and can be observed that it is irrelated to the value of $\beta$.
\begin{equation}\label{eq:theory-backward-reduction}
    \small
        1-\frac{10LN^3}{(12L-2)N^3}\in [0\%,16.67\%),~
        s.t.~L\in [1,+\infty).
\end{equation}

When $\beta \to 0$, the multi-exit networks degrade to a cascade of shallow networks with total training operations of $10LN^3$.
When $\beta \to 1$, the multi-exit networks degrade to backbone networks with multiple independent tasks, and the total training operations are $10LN^3$.
In conclusion, for a typical multi-exit network, employing \methodname~can save up to $16.67\%$ of operations in model training.

%% file: sections/05-Experiments.tex
\section{Experiments}

In this section, we conduct experiments to evaluate the accuracy and training efficiency improvement of multi-exit networks with \methodname.

\subsection{Model Accuracy Evaluation}\label{sec:accuracy-evaluation}

We use Deeply-Supervised Nets (DSN)~\cite{lee2015deeply}, BYOT~\cite{zhang2019your}, Gradient Equilibrium (GE)~\cite{li2019improved}, MSDNet~\cite{huang2018MSDNet}, PCGrad~\cite{yu2020gradient}, CAGrad~\cite{liu2021conflict},
Meta-GF~\cite{sun2022meta}, Block-Dependent Loss (BDL)~\cite{han2023improving}, WPN~\cite{han2022learning}, Nash-MTL~\cite{navon2022multi}, and DR-MGF~\cite{sun2023learning} as the baseline methods.
All accuracy results of baselines are cited from the original papers\footnote{The results of MSDNet are cited from \cite{li2019improved}, and the results of PCGrad and CAGrad for multi-exit network training are cited from \cite{sun2022meta}.}.
We use the same datasets and models used in the baselines and adopt \methodname~to the model training framework.
Specifically, the datasets are
Cifar100~\cite{krizhevsky2009learning} and ImageNet~\cite{deng2009imagenet}.
The data augmentation policies for Cifar100 and ImageNet refer to \cite{zhang2019your} and \cite{ESB}, respectively.
The models on these datasets are MSDNet~\cite{huang2018MSDNet} models and ResNet18~\cite{he2016deep} and their implementations in \cite{li2019improved} and \cite{zhang2019your} are utilized for reproducibility and fair comparison.
$\beta=0.5$ is used as the default setting for \methodname.
For Cifar100 dataset, we use the stochastic gradient descent (SGD) optimizer with a momentum of 0.9 for all models.
We train all models from scratch for a total of 300 epochs with an initial learning rate of 0.1 and drops to 0.01, 1e-3, and 1e-4, after 250, 280, and 295 epochs, a weight decay of $5e^{-4}$, and a batch size of 500.
For ImageNet dataset, we use the SGD optimizer with a batch size of 256, and train MSDNet models from scratch for a total of 90 epochs, with an initial learning rate of 0.1 and drops to 0.01, 1e-3, and 1e-4, after 70, 80, and 85 epochs.
Besides, we do not employ extra training tricks, such as knowledge distillation strategies~\cite{huang2018MSDNet,zhang2019your}, gradient selection strategies~\cite{yu2020gradient,sun2022meta,navon2022multi}, and weighted losses or samples~\cite{han2022learning,sun2023learning} in this evaluation.
The top-1 accuracy is used as the metric for accuracy evaluation for each of the exits.

\begin{table}[!t]
    \centering
    \caption{Comparison results of ResNet18 on Cifar100.}\vspace{-10pt}
    \label{tab:cifar100_results_on_resnet18}
    \setlength{\tabcolsep}{13pt} 
    \begin{tabular}{l|ccccc}
        \hline
        Methods                   & Exit1          & Exit2          & Exit3          & Exit4          & Ensemble       \\ \hline
        Params (M)                & 0.43           & 0.96           & 3.11           & 11.17          & -              \\
        FLOPs (M)                 & 169.62         & 300.22         & 431.05         & 559.75         & -              \\\hline
        DSN~\cite{lee2015deeply}  & 67.23          & 73.80          & 77.75          & 78.38          & 79.27          \\
        BYOT~\cite{zhang2019your} & 67.85          & 74.57          & 78.23          & 78.64          & 79.67          \\\hline
        \textbf{\methodname}      & \textbf{74.17} & \textbf{77.43} & \textbf{78.82} & \textbf{79.93} & \textbf{80.89} \\\hline
    \end{tabular}\vspace{-5pt}
    \flushleft\small{
        *Ensemble is the accuracy of ensembled logits from all tasks~\cite{zhang2019your}.
    }
    \vspace{-10pt}
\end{table}
\subsubsection{Results of ResNet18 on Cifar100.}\label{sec:results-of-resnet18-on-cifar100}
The experimental results for ResNet18 on Cifar100 dataset are shown in \tabref{tab:cifar100_results_on_resnet18}.
Compared with DSN~\cite{lee2015deeply} and BYOT~\cite{zhang2019your}, \methodname~achieves higher accuracy across all exits by mitigating gradient conflict and reusing the multi-scale features of different exits.
For early exits, such as exit 1 and exit 2, the accuracy improvements are up to $\textbf{6.94\%}$ and $\textbf{3.63\%}$, respectively.
The ensemble accuracy results of \methodname~also outperform the same counterparts from DSN and BYOT by $\textbf{1.62\%}$ and $\textbf{1.22\%}$, respectively.
The results show that applying \methodname~largely improves the accuracy of the exits, especially the accuracy of early exits.
This is because the early exits share all parameters with all exits, which are more susceptible to the impact of gradient conflict, thus their accuracy results often decrease more when compared to single-task models and deep tasks.
\methodname~mitigates the gradient conflict in training, leading to a significant improvement in the performance of early exits.

\begin{table}[!t]
    \centering
    \caption{Comparison results of MSDNet on Cifar100.}\vspace{-10pt}
    \label{tab:cifar100_results_on_msdnet}
    \setlength{\tabcolsep}{6.5pt} 
    \begin{tabular}{l|ccccccc}\hline
        Methods                               & Exit1          & Exit2          & Exit3          & Exit4          & Exit5         & Exit6          & Exit7          \\ \hline
        Params (M)                            & 0.90           & 1.84           & 2.80           & 3.76           & 4.92          & 6.10           & 7.36           \\
        FLOPs (M)                             & 56.43          & 101.00         & 155.31         & 198.10         & 249.53        & 298.05         & 340.64         \\\hline
        \small{MSDNet}~\cite{huang2018MSDNet} & 64.10          & 67.46          & 70.34          & 72.38          & 73.06         & 73.81          & 73.89          \\
        \small{GE}~\cite{li2019improved}      & 64.00          & 68.41          & 71.86          & 73.50          & 74.46         & 75.39          & 75.96          \\
        \small{CAGrad}~\cite{liu2021conflict} & 68.78          & 72.55          & 74.23          & 74.97          & 75.35         & 75.82          & 76.08          \\
        \small{PCGrad}~\cite{yu2020gradient}  & 67.06          & 71.37          & 74.86          & 75.78          & 76.25         & 76.95          & 76.71          \\
        \small{Meta-GF}~\cite{sun2022meta}    & 67.97          & 72.27          & 75.06          & 75.77          & 76.38         & 77.11          & 77.47          \\
        \small{BDL-1}~\cite{han2023improving} & 66.29          & 69.31          & 71.15          & 72.05          & 72.61         & 73.23          & 73.59          \\
        \small{BDL-2}~\cite{han2023improving} & 65.81          & 68.93          & 71.01          & 72.45          & 72.98         & 73.42          & 74.27          \\
        \small{WPN}~\cite{han2022learning}    & 62.26          & 67.18          & 70.53          & 73.10          & 74.80         & 76.05          & 76.31          \\
        \small{Nash-MTL}\cite{navon2022multi} & 64.14          & 69.23          & 72.64          & 74.89          & 75.32         & 75.88          & 76.75          \\
        \small{DR-MGF}~\cite{sun2023learning} & 67.82          & 72.45          & 74.77          & 75.77          & 76.56         & 76.90          & 77.01          \\
        \hline
        \textbf{\methodname}                  & \textbf{72.46} & \textbf{75.64} & \textbf{77.50} & \textbf{78.54} & \textbf{79.2} & \textbf{79.45} & \textbf{79.56} \\ \hline
    \end{tabular}
    \vspace{-10pt}
\end{table}

\subsubsection{Results of MSDNet on Cifar100.}\label{sec:results-of-msdnet-on-cifar100}

As shown in \tabref{tab:cifar100_results_on_msdnet}, the
results demonstrate that \methodname~consistently outperforms all baselines across all exits of MSDNet, and significantly improves the overall performance.
Compared with the baselines, the average accuracy improvements of \methodname~on the 7 exits are $\textbf{6.64\%}$, $\textbf{5.72\%}$,$\textbf{4.86\%}$, $\textbf{4.47\%}$, $\textbf{4.42\%}$, $\textbf{3.99\%}$, and $\textbf{3.76\%}$, respectively.
It can be observed that the accuracy improvements of early exits are more significant than those of deep ones, which is because of the mitigation of the gradient conflict that improves the performance of early tasks, as described in \secref{sec:results-of-resnet18-on-cifar100}.
The average accuracy improvement of \methodname~across all tasks and all baselines is $\textbf{4.84\%}$.

\begin{table}[!t]
    \centering
    \caption{Comparison results of MSDNet on ImageNet.
    }
    \vspace{-10pt}
    \label{tab:imagenet_results_on_msdnet}
    \setlength{\tabcolsep}{9.8pt} 
    \begin{tabular}{l|cccccc}\hline
        Methods                       & Exit1          & Exit2          & Exit3          & Exit4          & Exit5          & \small{Average} \\ \hline
        Params (M)                    & 4.24           & 8.77           & 13.07          & 16.75          & 23.96          & -               \\
        FLOPs (G)                     & 0.34           & 0.69           & 1.01           & 1.25           & 1.36           & -               \\\hline
        MSDNet~\cite{huang2018MSDNet} & 56.64          & 65.14          & 68.42          & 69.77          & 71.34          & 66.26           \\
        GE~\cite{li2019improved}      & 57.28          & 66.22          & 70.24          & 71.71          & 72.43          & 67.58           \\
        CAGrad~\cite{liu2021conflict} & 58.37          & 64.21          & 66.88          & 68.22          & 69.42          & 65.42           \\
        PCGrad~\cite{yu2020gradient}  & 57.62          & 64.87          & 68.93          & 71.05          & 72.45          & 66.98           \\
        Mate-GF~\cite{sun2022meta}    & 57.43          & 64.82          & 69.08          & 71.67          & \textbf{73.27} & 67.25           \\\hline
        \textbf{\methodname}          & \textbf{61.80} & \textbf{68.03} & \textbf{70.75} & \textbf{71.79} & 72.88          & \textbf{69.05}  \\\hline
    \end{tabular}\vspace{-5pt}
    \flushleft\small{
        Average is the average accuracy of 5 tasks~\cite{sun2022meta}.
    }\vspace{-10pt}
\end{table}
\subsubsection{Results of MSDNet on ImageNet.}
The
results are shown in \tabref{tab:imagenet_results_on_msdnet}.
Results show that
\methodname~outperforms all the baselines at the early exits even without extra training tricks, and the average accuracy improvements of \methodname~on the 5 tasks are
$\textbf{4.33\%}$, $\textbf{2.98\%}$, $\textbf{2.04\%}$, $\textbf{1.31\%}$, $\textbf{1.10\%}$,
respectively.
Similar to the experiments on Cifar100, the accuracy improvements on early exits are more significant than those of deep ones, which further demonstrates the effectiveness of \methodname~on conflict mitigation.
The average accuracy achieved by \methodname~is $\textbf{69.05\%}$, which significantly exceeds baselines, and the average accuracy improvement achieves $\textbf{2.35\%}$.
One exception happens at the fifth exit, the accuracy of \methodname~is slightly lower than that of Mate-GF~\cite{sun2022meta}. 
The main reason is that \methodname~relies on the over-parameterization characteristics of models for higher accuracy while the model size of MSDNet is relatively small, especially when training on the ImageNet dataset.
It hinders the accuracy improvement of \methodname~for the later tasks that have greater differences in model capacity compared to early tasks.
Besides, Mate-GF employed a cautious while time-consuming gradient merge method to achieve high accuracy, while \methodname~do not employ any additional methods for high training efficiency, which is another cause for the results.

\begin{wrapfigure}{r}{0.5\textwidth}
    \centering
    \vspace{-25pt}
    \includegraphics[width=0.5\textwidth]{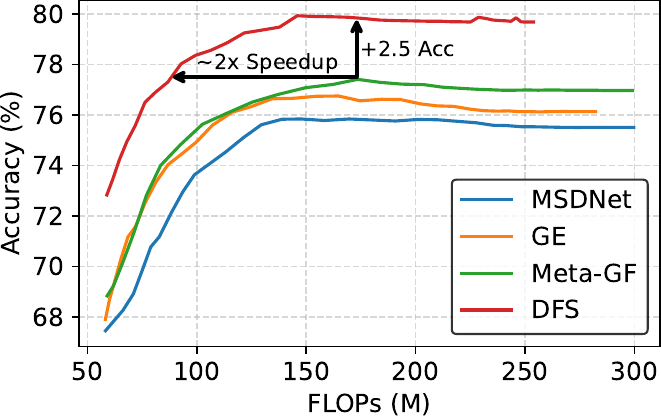}
    \vspace{-20pt}
    \caption{Accuracy of budgeted batch classification. 
    The X-axis is the average computational budget per image for MSDNet model on the Cifar100,
    and Y-axis top-1 accuracy.
    }
    \label{fig:budgetedbatchclassification}
    \vspace{-30pt}
\end{wrapfigure}

\vspace{-10pt}
\subsubsection{Budgeted Batch Classification.}
We evaluate the achieved classification accuracy of the model within a given computational budget~\cite{li2019improved,sun2022meta}.
We conduct experiments on Cifar100 dataset using MSDNet~\cite{huang2018MSDNet} model.
The baseline methods include MSDNet~\cite{huang2018MSDNet}, GE~\cite{li2019improved}, and Meta-GF~\cite{sun2022meta}.
The results are shown in \figref{fig:budgetedbatchclassification}.
The results show that \methodname~consistently outperforms all baselines across all budgets.
With an average budget of 170MFLOPs, DFS achieves an accuracy of ${\sim}79.5\%$, which is ${\boldsymbol{\sim}}\textbf{2.5\%}$ higher than that of Meta-GF with the same budget.
DFS uses ${\boldsymbol{\sim}}\textbf{2}{\boldsymbol{\times}}$ fewer FLOPs to achieve the same classification accuracy compared to Meta-GF, which is the best among all baselines.

\subsection{Training Efficiency Validation}\label{sec:training-efficiency-validation}

In this experiment, we evaluate the training efficiency of \methodname.
Specifically, the computation reduction during the backpropagation when adopting \methodname~contributes to efficient training of multi-exit networks.
We select $\beta=0.5$ as the default setting for \methodname.

We adopt the
VGG7-64 in~\cite{ESB} with 6 convolution layers and train it on Cifar100~\cite{krizhevsky2009learning} dataset, and VGG16~\cite{simonyan2014vgg} with 13 convolution layers and train it on ImageNet~\cite{krizhevsky2012imagenet} dataset to evaluate the training efficiency improvement from \methodname.
Both models employ a Batch Normalization (BN)~\cite{ioffe2015batch} layer after each convolution layer and the early classifiers are added after the BNs.
Each classifier only contains a Global Average Pooling (GAP)~\cite{lin2013network} layer and a Fully-Connected (FC) layer for classification.
The baseline methods include DSN~\cite{lee2015deeply} and BYOT~\cite{zhang2019your}.
DSN applies the vanilla training in~\cref{sec:implementation} which optimizes all losses with the same weight and updates the trainable parameters.
BYOT introduces self-distillation supervision based on DSN, which employs the outputs and logits of the final exit to teach the early tasks.
We use the latency for training one batch of samples and Frames Per Second (FPS) on the same devices to evaluate the training efficiency.
The overall training latency is divided into forward latency, backward latency, and other latency for detailed comparison. The other latency includes data loading, loss computation, parameter update, etc.
The lower latency and higher FPS stand for a higher training efficiency.
For a fair comparison, the results are collected with different batch sizes.
We train VGG7-64 on CPU (2$\times$ 20-core Intel Xeon Gold 6248) and VGG16 on GPU (1$\times$ NVIDIA RTX A6000), both are trained with 100 batches of data inputs.

\begin{figure}[!t]
    \centering
    \subfloat[Results of VGG7-64 over Cifar100 on CPU]{
        \includegraphics[width=0.475\textwidth]{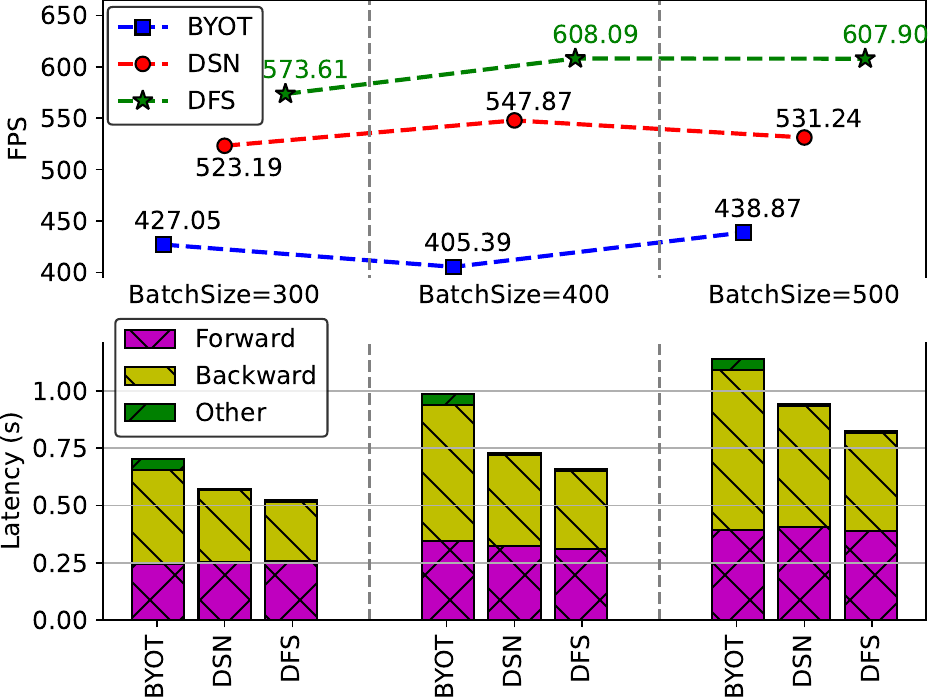}
    }
    \subfloat[Results of VGG16 over ImageNet on GPU]{
        \includegraphics[width=0.475\textwidth]{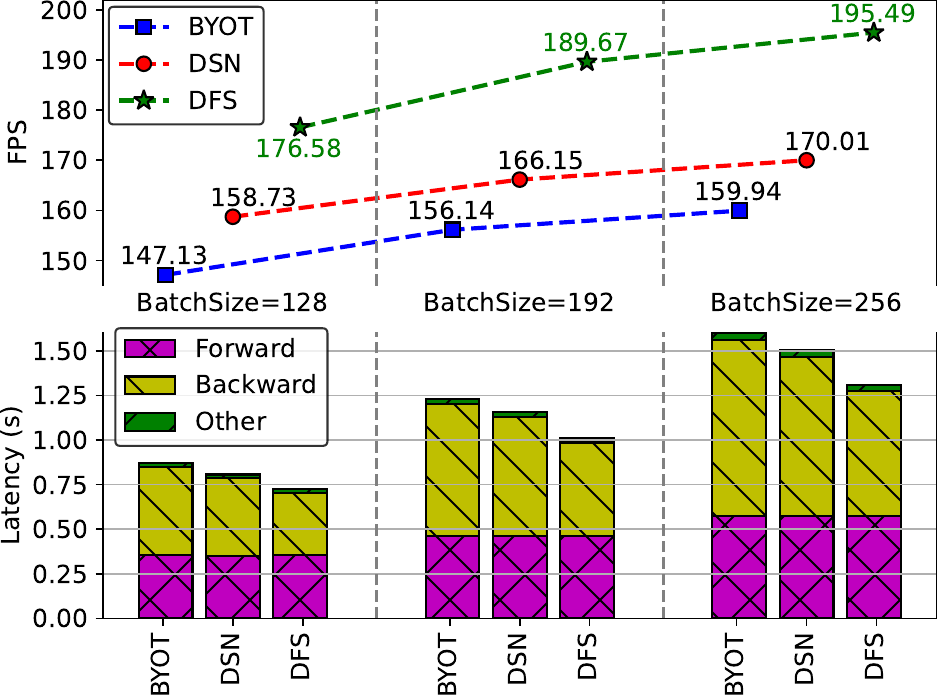}
    }
    \vspace{-5pt}
    \caption{Training efficiency comparison.
        Y1-axis is the average latency (divided into forward, backward, and other parts) for one training step, Y2-axis is the average FPS, and X-axis is the results of DSN, BYOT, and \methodname~under different batch sizes.}
    \label{fig:training-efficiency-vgg-on-cifar100-on-CPU}
    \vspace{-5pt}
\end{figure}

\begin{table}[!t]
    \centering
    \caption{The accuracy comparison of multiple tasks in the VGG7-64 model with different coefficient values.
    }\vspace{-10pt}
    \label{tab:accuracy-of-vgg-on-cifar100-across-beta}
    \setlength{\tabcolsep}{9.8pt} 
    \begin{tabular}{l|cccccc}
        \hline
        Methods                 & Exit1          & Exit2          & Exit3          & Exit4          & Exit5          & Exit6          \\ \hline
        Params (K)              & 8.32           & 45.31          & 125.70         & 273.41         & 581.63         & 1171.97        \\
        FLOPs (M)               & 2.04           & 40.05          & 59.06          & 96.94          & 115.89         & 153.71         \\\hline
        \methodname~$\beta=0.1$ & \textbf{22.81} & \textbf{41.56} & 53.93          & 58.58          & 63.87          & 65.46          \\
        \methodname~$\beta=0.3$ & 21.23          & 40.34          & \textbf{55.53} & 61.31          & 67.11          & 69.44          \\
        \methodname~$\beta=0.5$ & 20.77          & 39.21          & 55.31          & \textbf{61.83} & 68.61          & 70.56          \\
        \methodname~$\beta=0.7$ & 19.62          & 36.42          & 52.96          & 60.95          & \textbf{69.04} & 72.17          \\
        \methodname~$\beta=0.9$ & 17.88          & 32.24          & 49.69          & 57.12          & 68.13          & \textbf{73.06} \\ \hline
    \end{tabular}
    \vspace{-10pt}
\end{table}

The experimental results are shown in \figref{fig:training-efficiency-vgg-on-cifar100-on-CPU}.
For different batch sizes and devices, all the DSN, BYOT, and \methodname~have similar forward latency, since their forward operations are the same.
For the backward latency, \methodname~has shorter latency compared with DSN and BYOT across all batch sizes and devices mainly because
\methodname~reduces the required number of operations for backward computation.
BYOT uses knowledge distillation and intermediate feature distillation to improve model accuracy, resulting in much higher backward latency and other latency than DSN and \methodname.

The FPS results of VGG7-64 with \methodname~outperforms DSN and BYOT by up to $\textbf{14.43\%}$ (${76.67 fps}$) and $\textbf{50.00\%}$ (${202.70 fps}$), and the average FPS improvements across batch sizes are $\textbf{11.69\%}$ (${62.43 fps}$) and $\textbf{40.77\%}$ (${172.77 fps}$), respectively.
On the GPU device with VGG16 as backbone, the training FPS results of \methodname~excced DSN and BYOT by up to $\textbf{14.99\%}$ (${25.48 fps}$) and $\textbf{22.23\%}$ ($35.55 fps$), and the average FPS improvements across batch sizes are $\textbf{13.51\%}$ ($22.28 fps$) and $\textbf{21.27\%}$ ($32.84 fps$), respectively.
The results demonstrate that \methodname~reduces the backward latency significantly during model training and improves FPS across different models, batch sizes, and devices.
The actual average FPS improvements obtained through experiments are $\textbf{11.69\%}$ and $\textbf{13.51\%}$, both of them are consistent with the theoretical FPS improvements of VGG7-64 and VGG16 according to \cref{eq:theory-backward-reduction}, which are $\textbf{14.29\%}$ with $L=6$ (convolutional layers of VGG7-64) and $\textbf{15.58\%}$ with $L=13$ (convolutional layers of VGG16).

\begin{figure}[!t]
    \centering
    \includegraphics[width=1\textwidth]{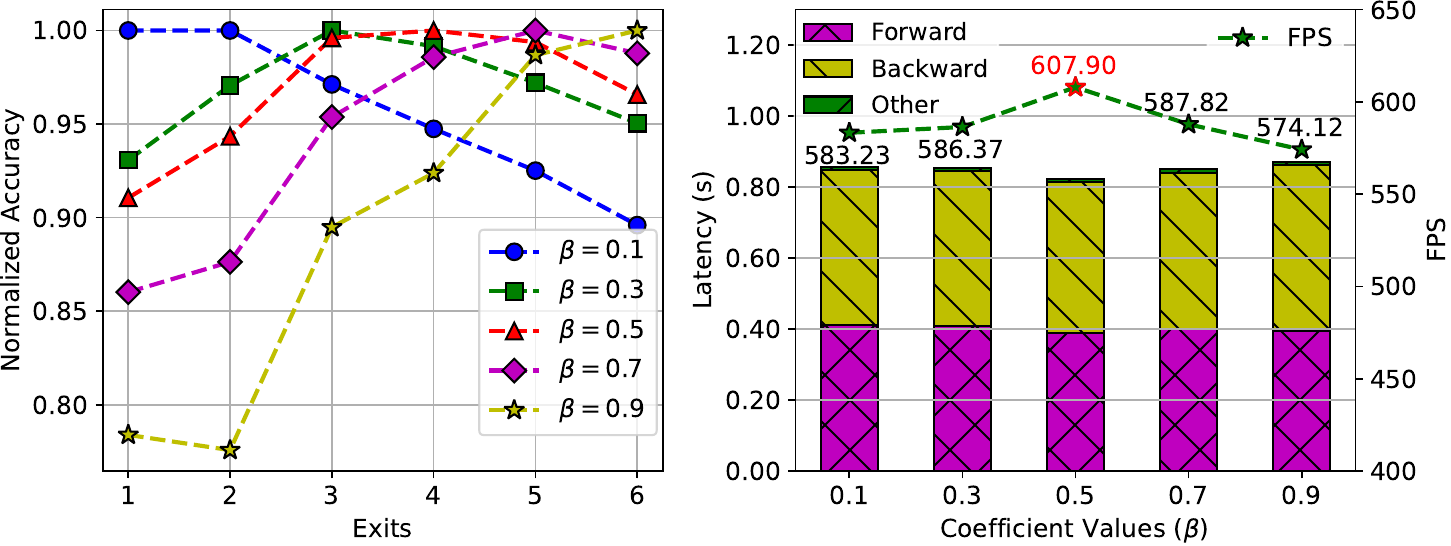}
    \vspace{-20pt}
    \caption{The impact of the coefficient on task accuracy and training efficiency.
        The left side shows the normalized accuracy of \methodname~with different $\beta$.
        It focuses on the sort instead of the absolute value of accuracy results on each task.
        The right side is the training efficiency of VGG7-64 using \methodname~with different $\beta$.
        Y1-axis is the average latency and y2-axis is the average FPS metric.
    }
    \label{fig:training-efficiency-of-vgg-across-coefficient-values}
    \vspace{-0pt}
\end{figure}
\subsection{Impact of Coefficient}\label{sec:coefficient-selection}

We evaluate the impact of the coefficient on the accuracy and training efficiency of \methodname~in this experiment.
We select VGG7-64 on Cifar100 with batch size 500 to evaluate the impact of $\beta$ on the model accuracy and training latency (FPS).

The accuracy of different exits with different coefficients is shown in \tabref{tab:accuracy-of-vgg-on-cifar100-across-beta} and the normalized accuracy with training latency (FPS) is shown in \figref{fig:training-efficiency-of-vgg-across-coefficient-values}.
The normalized accuracy is normalized to the maximum accuracy of different methods on each task to show the impact of $\beta$.
The results show that
a lower coefficient value results in a higher accuracy for early exits while a lower accuracy for later exits.
On the contrary, the higher $\beta$ improves the accuracy of later exits while decreasing that of early exits.
In summary, \textbf{lower $\beta$ contributes to early exits while higher $\beta$ benefits later exits}.
This is because the lower $\beta$ assigns more exit-specific features and decreases the shared features.
When decreasing $\beta$, the number of trainable weights
for later exits decreases while the number of trainable weights for early tasks increases.
For exits with different depths, the best coefficient values are different.
As shown in \cref{fig:training-efficiency-of-vgg-across-coefficient-values},
an optimal coefficient value improves the training efficiency of multi-exit networks.
According to \cref{eq:backward_with_dns},
increasing $\beta$ will increase the portion of shared weights, thus increasing the computational load of $g_{w_i^+}$, while decreasing $\beta$
increases the computational load for $g_{w_i^-}$.
In our experiments,
$\beta=0.5$ offers a balanced computational load among multiple tasks and leads to higher training efficiency.

%% file: sections/06-Conclusion.tex
\section{Conclusion}
In this paper, we proposed a novel method \methodname~to mitigate the gradient conflict in the training of multi-exit networks.
Specifically,
\methodname~consists of feature partitioning and feature referencing techniques.
The former decouples the shared weights among multiple exits and mitigates the gradient conflict in model training and
ensures an end-to-end joint optimization for each exit.
The latter reuses the multi-scale features among exits and enhances the overall performance of multi-exit networks.
\methodname~improves the accuracy by up to 6.94\% of the individual exits and up to 3.63\% for the overall ensembled model accuracy.
In addition, we provide a detailed analysis of the operation reduction for backpropagation by applying \methodname~and the experimental results show that \methodname~saves up to 50.00\% training time in the training of multi-exit networks and still provides improved model accuracy.

%% file: sections/07-Acknowledge.tex
\section{Acknowledgments}
This work is partially supported by the China Postdoctoral Science Foundation (2022M721707),
the National Natural Science Foundation (62272248),
the Natural Science Foundation of Tianjin (23JCZDJC01010, 23JCQNJC00010),
the National Natural Science Foundation (62372253),
the Natural Science Foundation of Tianjin Fund (23JCYBJC00010),
the CCF-Baidu Open Fund (CCF-Baidu202310),
the Open Project Fund of State Key Laboratory of Computer Architecture, ICT, CAS (CARCHB202016),
the Innovation Fund of Qiyuan Lab (2022-JCJ0-LA-001-068),
Open Fund of Civil Aviation Smart Airport Theory and System Key laboratory, Civil Aviation University of China (SATS202303),
the Key Program for International Cooperation of Ministry of Science and Technology, China (2024YFE0100700).